\documentclass[11pt,a4paper]{article}
\usepackage[hyperref]{emnlp-ijcnlp-2019}
\usepackage{times}
\usepackage{latexsym}

\usepackage{url}
\usepackage{enumitem}
\usepackage{xspace}
\usepackage{booktabs}

\usepackage{multirow}
\usepackage{graphicx}
\usepackage{siunitx}
\usepackage{soul}
\usepackage{tikz}

\usepackage{amsmath}
\usepackage{amssymb}
\usepackage[normalem]{ulem}

\aclfinalcopy % Uncomment this line for the final submission

\newcommand\fever{FEVER\xspace}
\newcommand\hpqa{\textsc{HotpotQA}\xspace}

\title{Revealing the Importance of Semantic Retrieval \\ for Machine Reading at Scale}

\author{Yixin Nie \;\;\;\;\;\; Songhe Wang \;\;\;\;\;\; Mohit Bansal\\
  UNC Chapel Hill \\
  {\tt \{yixin1, songhe17, mbansal\}@cs.unc.edu} \\
}

\date{}

\begin{document}
\maketitle
\begin{abstract}
  Machine Reading at Scale (MRS) is a challenging task in which a system is given an input query and is asked to produce a precise output by ``reading'' information from a large knowledge base. The task has gained popularity with its natural combination of information retrieval (IR) and machine comprehension (MC).
  Advancements in representation learning have led to separated progress in both IR and MC; however, very few studies have examined the relationship and combined design of retrieval and comprehension at different levels of granularity, for development of MRS systems.
  In this work, we give general guidelines on system design for MRS by proposing a simple yet effective pipeline system with special consideration on hierarchical semantic retrieval at both paragraph and sentence level, and their potential effects on the downstream task.
  The system is evaluated on both fact verification and open-domain multi-hop QA, achieving state-of-the-art results on the leaderboard test sets of both \fever and \hpqa.
  To further demonstrate the importance of semantic retrieval, we present ablation and analysis studies to quantify the contribution of neural retrieval modules at both paragraph-level and sentence-level, and illustrate that intermediate semantic retrieval modules are vital for not only effectively filtering upstream information and thus saving downstream computation, but also for shaping upstream data distribution and providing better data for downstream modeling.\footnote{Code/data made publicly available at: \url{https://github.com/easonnie/semanticRetrievalMRS}}
\end{abstract}

\section{Introduction}

Extracting external textual knowledge for machine comprehensive systems has long been an important yet challenging problem. Success requires not only precise retrieval of the relevant information sparsely restored in a large knowledge source but also a deep understanding of both the selected knowledge and the input query to give the corresponding output. Initiated by~\newcite{chen2017drqa}, the task was termed as Machine Reading at Scale (MRS), seeking to provide a challenging situation where machines are required to do both semantic retrieval and comprehension at different levels of granularity for the final downstream task.

Progress on MRS has been made by improving individual IR or comprehension sub-modules with recent advancements on representative learning~\cite{peters2018ELMo, radford2018gpt_1, devlin2018bert}.
However, partially due to the lack of annotated data for intermediate retrieval in an MRS setting, the evaluations were done mainly on the final downstream task and with much less consideration on the intermediate retrieval performance. This led to the convention that upstream retrieval modules mostly focus on getting better coverage of the downstream information such that the upper-bound of the downstream score can be improved, rather than finding more exact information. This convention is misaligned with the nature of MRS where equal effort should be put in emphasizing the models' joint performance and optimizing the relationship between the semantic retrieval and the downstream comprehension sub-tasks. 

Hence, to shed light on the importance of semantic retrieval for downstream comprehension tasks, we start by establishing a simple yet effective hierarchical pipeline system for MRS using Wikipedia as the external knowledge source. 
The system is composed of a term-based retrieval module, two neural modules for both paragraph-level retrieval and sentence-level retrieval, and a neural downstream task module.
We evaluated the system on two recent large-scale open domain benchmarks for fact verification and multi-hop QA, namely \fever~\cite{Thorne18Fever} and \hpqa~\cite{yang2018hotpotqa}, in which retrieval performance can also be evaluated accurately since intermediate annotations on evidences are provided.
Our system achieves the start-of-the-art results with 45.32\% for answer EM and 25.14\% joint EM on \hpqa (8\% absolute improvement on answer EM and doubling the joint EM over the previous best results) and with 67.26\% on \fever score (3\% absolute improvement over previously published systems).

We then provide empirical studies to validate design decisions. Specifically, we prove the necessity of both paragraph-level retrieval and sentence-level retrieval for maintaining good performance, and further illustrate that a better semantic retrieval module not only is beneficial to achieving high recall and keeping high upper bound for downstream task, but also plays an important role in shaping the downstream data distribution and providing more relevant and high-quality data for downstream sub-module training and inference. These mechanisms are vital for a good MRS system on both QA and fact verification.

\section{Related Work}

\noindent\textbf{Machine Reading at Scale}
First proposed and formalized in \newcite{chen2017drqa}, MRS has gained popularity with increasing amount of work on both dataset collection~\cite{Joshi2017trivia_qa, welbl2018constructing} and MRS model developments~\cite{wang2018r3, clark2017simple, htut2018training}. In some previous work~\cite{lee2018ranking}, paragraph-level retrieval modules were mainly for improving the recall of required information, while in some other works~\cite{yang2018hotpotqa}, sentence-level retrieval modules were merely for solving the auxiliary sentence selection task. In our work, we focus on revealing the relationship between semantic retrieval at different granularity levels and the downstream comprehension task. To the best of our knowledge, we are the first to apply and optimize neural semantic retrieval at both paragraph and sentence levels for MRS.

\noindent\textbf{Automatic Fact Checking:} Recent work \cite{thorne2018automated_survey} formalized the task of automatic fact checking from the viewpoint of machine learning and NLP. The release of \fever~\cite{Thorne18Fever} stimulates many recent developments~\cite{nie2019combining, yoneda2018ucl_2nd, hanselowski2018ukp_3rd} on data-driven neural networks for automatic fact checking. We consider the task also as MRS because they share almost the same setup except that the downstream task is verification or natural language inference (NLI) rather than QA.

\noindent\textbf{Information Retrieval}
Success in deep neural networks inspires their application to information retrieval (IR) tasks~\cite{huang2013learning,guo2016deep,mitra2017learning,dehghani2017neural}.
In typical IR settings, systems are required to retrieve and rank~\cite{nguyen2016ms_marco} elements from a collection of documents based on their relevance to the query. This setting might be very different from the retrieval in MRS where systems are asked to select facts needed to answer a question or verify a statement. We refer the retrieval in MRS as \textit{Semantic Retrieval} since it emphasizes on semantic understanding.

\begin{figure*}[t]
\centering
\includegraphics[width=0.95\textwidth]{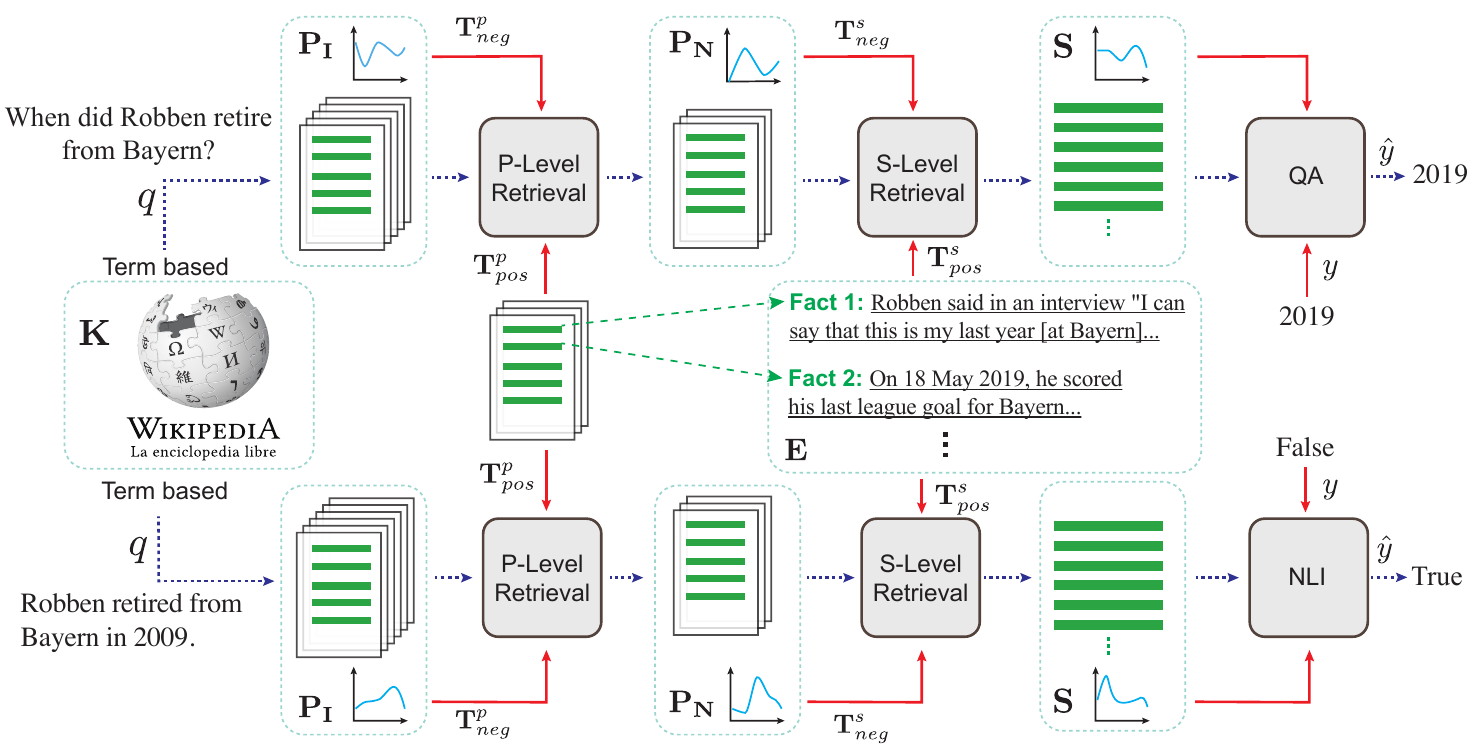}
\vspace{-7pt}
\caption{System Overview: blue dotted arrows indicate the inference flow and the red solid arrows indicate the training flow. Grey rounded rectangles are neural modules with different functionality. The two retrieval modules were trained with all positive examples from annotated ground truth set and negative examples sampled from the direct upstream modules. Thus, the distribution of negative examples is subjective to the quality of the upstream module.}
\label{fig:fig_system_overview}
\vspace{-10pt}
\end{figure*}

\section{Method}
\label{sec:system_setup}
In previous works, an MRS system can be complicated with different sub-components processing different retrieval and comprehension sub-tasks at different levels of granularity, and with some sub-components intertwined. For interpretability considerations, we used a unified pipeline setup. The overview of the system is in Fig.~\ref{fig:fig_system_overview}.

To be specific, we formulate the MRS system as a function that maps an input tuple $(q, \mathbf{K})$ to an output tuple $(\hat{y}, \mathbf{S})$ where $q$ indicates the input query, $\mathbf{K}$ is the textual KB, $\hat{y}$ is the output prediction, and $\mathbf{S}$ is selected supporting sentences from Wikipedia. Let $\mathbf{E}$ denotes a set of necessary evidences or facts selected from $\mathbf{K}$ for the prediction. For a QA task, $q$ is the input question and $\hat{y}$ is the predicted answer. For a verification task, $q$ is the input claim and $\hat{y}$ is the predicted truthfulness of the input claim. For all tasks, $\mathbf{K}$ is Wikipedia.

The system procedure is listed below:

\noindent\textbf{(1) Term-Based Retrieval:} To begin with, we used a combination of the TF-IDF method and a rule-based keyword matching method\footnote{Details of term-based retrieval are in Appendix.} to narrow the scope from whole Wikipedia down to a set of related paragraphs; this is a standard procedure in MRS~\cite{chen2017drqa, lee2018ranking, nie2019combining}. The focus of this step is to efficiently select a candidate set $\mathbf{P_I}$ that can cover the information as much as possible ($\mathbf{P_I} \subset \mathbf{K}$) while keeping the size of the set acceptable enough for downstream processing.

\noindent\textbf{(2) Paragraph-Level Neural Retrieval:} After obtaining the initial set, we compare each paragraph in $\mathbf{P_I}$ with the input query $q$ using a neural model (which will be explained later in Sec~\ref{sec:model_train}).
The outputs of the neural model are treated as the relatedness score between the input query and the paragraphs. The scores will be used to sort all the upstream paragraphs. Then, $\mathbf{P_I}$ will be narrowed to a new set $\mathbf{P_N}$ ($\mathbf{P_N} \subset \mathbf{P_I}$) by selecting top $k_p$ paragraphs having relatedness score higher than some threshold value $h_p$ (going out from the P-Level grey box in Fig.~\ref{fig:fig_system_overview}). $k_p$ and $h_p$ would be chosen by keeping a good balance between the recall and precision of the paragraph retrieval.

\noindent\textbf{(3) Sentence-Level Neural Retrieval:} Next, we select the evidence at the sentence-level by decomposing all the paragraphs in $\mathbf{P_N}$ into sentences. Similarly, each sentence is compared with the query using a neural model (see details in Sec~\ref{sec:model_train}) and obtain a set of sentences $\mathbf{S} \subset \mathbf{P_N}$ for the downstream task by choosing top $k_s$ sentences with output scores higher than some threshold $h_s$ (S-Level grey box in Fig.~\ref{fig:fig_system_overview}). During evaluation, $\mathbf{S}$ is often evaluated against some ground truth sentence set denoted as $\mathbf{E}$.

\noindent\textbf{(4) Downstream Modeling:}
At the final step, we simply applied task-specific neural models (e.g., QA and NLI) on the concatenation of all the sentences in $\mathbf{S}$ and the query, obtaining the final output $\hat{y}$.

In some experiments, we modified the setup for certain analysis or ablation purposes which will be explained individually in Sec~\ref{sec:analysis}.

\subsection{Modeling and Training}
\label{sec:model_train}
Throughout all our experiments, we used \textbf{BERT-Base}~\cite{devlin2018bert} to provide the state-of-the-art contextualized modeling of the input text.\footnote{We used the pytorch BERT implementation in \url{https://github.com/huggingface/pytorch-pretrained-BERT}.}

\vspace{5pt}\textbf{Semantic Retrieval:} We treated the neural semantic retrieval at both the paragraph and sentence level as binary classification problems with models' parameters updated by minimizing binary cross entropy loss. To be specific, we fed the query and context into BERT as:
\begin{equation*}
\begin{small}
    [\mathit{CLS}] \text{ } \mathit{Query} \text{ }  [\mathit{SEP}] \text{ } \mathit{Context} \text{ } [\mathit{SEP}]
\end{small}
\end{equation*}
We applied an affine layer and sigmoid activation on the last layer output of the [$\mathit{CLS}$] token which is a scalar value. The parameters were updated with the objective function:
\begin{align*}
\mathcal{J}_{retri} = 
-\sum_{i \in \mathbf{T}^{p/s}_{pos}} \log(\hat{p}_i) -\sum_{i \in \mathbf{T}^{p/s}_{neg}} \log(1-\hat{p}_i)
\end{align*}
where $\hat{p}_i$ is the output of the model, $\mathbf{T}^{p/s}_{pos}$ is the positive set and $\mathbf{T}^{p/s}_{neg}$ is the negative set. As shown in Fig.~\ref{fig:fig_system_overview}, at sentence level, ground-truth sentences were served as positive examples while other sentences from upstream retrieved set were served as negative examples. Similarly at the paragraph-level, paragraphs having any ground-truth sentence were used as positive examples and other paragraphs from the upstream term-based retrieval processes were used as negative examples.

\vspace{5pt}\textbf{QA:} We followed \newcite{devlin2018bert} for QA span prediction modeling. To correctly handle yes-or-no questions in \hpqa, we fed the two additional ``$\mathit{yes}$" and ``$\mathit{no}$" tokens between [$\mathit{CLS}$] and the $Query$ as:
\begin{equation*}
\begin{small}
    [\mathit{CLS}] \text{ } \mathit{yes} \text{ } \mathit{no} \text{ } \mathit{Query} \text{ }  [\mathit{SEP}] \text{ } \mathit{Context} \text{ } [\mathit{SEP}]
\end{small}
\end{equation*}
where the supervision was given to the second or the third token when the answer is ``yes" or ``no", such that they can compete with all other predicted spans.
The parameters of the neural QA model were trained to maximize the log probabilities of the true start and end indexes as:
\begin{align*}
\mathcal{J}_{qa} = 
-\sum_{i} \big[ \log(\hat{y}^s_i) + \log(\hat{y}^e_i) \big]
\end{align*}
where $\hat{y}^s_i$ and $\hat{y}^e_i$ are the predicted probability on the ground-truth start and end position for the $i$th example, respectively. It is worth noting that we used ground truth supporting sentences plus some other sentences sampled from upstream retrieved set as the context for training the QA module such that it will adapt to the upstream data distribution during inference.

\vspace{5pt}\textbf{Fact Verification:}
Following~\newcite{Thorne18Fever}, we formulate downstream fact verification as the 3-way natural language inference (NLI) classification problem~\cite{maccartney2009NLI, snli:emnlp2015} and train the model with 3-way cross entropy loss. The input format is the same as that of semantic retrieval and the objective is $\mathcal{J}_{ver} = 
-\sum_{i} \mathbf{y}_i \cdot \log(\hat{\mathbf{y}}_i)$, where $\hat{\mathbf{y}}_i \in \mathbf{R^3}$ denotes the model's output for the three verification labels, and $\mathbf{y}_i$ is a one-hot embedding for the ground-truth label.
For verifiable queries, we used ground truth evidential sentences plus some other sentences sampled from upstream retrieved set as new evidential context for NLI. For non-verifiable queries, we only used sentences sampled from upstream retrieved set as context because those queries are not associated with ground truth evidential sentences. This detail is important for the model to identify non-verifiable queries and will be explained more in Sec~\ref{sec:analysis}.
Additional training details and hyper-parameter selections are in the Appendix (Sec.~\ref{sec:training_appendix}; Table~\ref{tab:hyper_parameter_pipeline_system}).

It is worth noting that each sub-module in the system relies on its preceding sub-module to provide data both for training and inference. This means that there will be upstream data distribution misalignment if we trained the sub-module in isolation without considering the properties of its precedent upstream module. The problem is similar to the concept of internal covariate shift~\cite{ioffe2015batchnorm}, where the distribution of each layer's inputs changes inside a neural network. Therefore, it makes sense to study this issue in a joint MRS setting rather than a typical supervised learning setting where training and test data tend to be fixed and modules being isolated.
We release our code and the organized data both for reproducibility and providing an off-the-shelf testbed to facilitate future research on MRS.

\section{Experimental Setup}
MRS requires a system not only to retrieve relevant content from textual KBs but also to poccess enough understanding ability to solve the downstream task. To understand the impact or importance of semantic retrieval on the downstream comprehension, we established a unified experimental setup that involves two different downstream tasks, i.e., multi-hop QA and fact verification.

\subsection{Tasks and Datasets}
\noindent\textbf{\hpqa:} This dataset is a recent large-scale QA dataset that brings in new features: (1) the questions require finding and reasoning over multiple documents; (2) the questions are diverse and not limited to pre-existing KBs; (3) it offers a new comparison question type~\cite{yang2018hotpotqa}. We experimented our system on \hpqa in the fullwiki setting, where a system must find the answer to a question in the scope of the entire Wikipedia, an ideal MRS setup. The sizes of the train, dev and test split are 90,564, 7,405, and 7,405.
More importantly, \hpqa also provides human-annotated sentence-level supporting facts that are needed to answer each question. Those intermediate annotations enable evaluation on models' joint ability on both fact retrieval and answer span prediction, facilitating our direct analysis on the explainable predictions and its relations with the upstream retrieval.

\noindent\textbf{FEVER:} The Fact Extraction and VERification dataset~\cite{Thorne18Fever} is a recent dataset collected to facilitate the automatic fact checking. The work also proposes a benchmark task in which given an arbitrary input claim, candidate systems are asked to select evidential sentences from Wikipedia and label the claim as either \textsc{Support}, \textsc{Refute}, or \textsc{Not Enough Info}, if the claim can be verified to be true, false, or non-verifiable, respectively, based on the evidence. The sizes of the train, dev and test split are 145,449, 19,998, and 9,998. Similar to \hpqa, the dataset provides annotated sentence-level facts needed for the verification. These intermediate annotations could provide an accurate evaluation on the results of semantic retrieval and thus suits well for the analysis on the effects of retrieval module on downstream verification.

As in~\newcite{chen2017drqa}, we use Wikipedia as our unique knowledge base because it is a comprehensive and self-evolving information source often used to facilitate intelligent systems. Moreover, as Wikipedia is the source for both \hpqa and \fever, it helps standardize any further analysis of the effects of semantic retrieval on the two different downstream tasks. 

\subsection{Metrics}
Following~\newcite{Thorne18Fever, yang2018hotpotqa}, we used annotated sentence-level facts to calculate the F1, Precision and Recall scores for evaluating sentence-level retrieval. Similarly, we labeled all the paragraphs that contain any ground truth fact as ground truth paragraphs and used the same three metrics for paragraph-level retrieval evaluation. For \hpqa, following \newcite{yang2018hotpotqa}, we used exact match (EM) and F1 metrics for QA span prediction evaluation, and used the joint EM and F1 to evaluate models' joint performance on both retrieval and QA. The joint EM and F1 are calculated as: $P_j = P_a \cdot P_s; R_j = R_a \cdot R_s; F_j = \frac{2P_j \cdot R_j}{P_j + R_j}; \text{EM}_j = \text{EM}_a \cdot \text{EM}_s$, where $P$, $R$, and $\text{EM}$ denote precision, recall and EM; the subscript $a$ and $s$ indicate that the scores are for answer span and supporting facts. 

For the \fever task, following \newcite{Thorne18Fever}, we used the Label Accuracy for evaluating downstream verification and the Fever Score for joint performance. Fever score will award one point for each example with the correct predicted label only if all ground truth facts were contained in the predicted facts set with at most 5 elements. We also used Oracle Score for the two retrieval modules. The scores were proposed in \newcite{nie2019combining} and indicate the upperbound of final FEVER Score at one intermediate layer assuming all downstream modules are perfect. 
All scores are averaged over examples in the whole evaluation set.

%------------------------------------------------
%------------------------------------------------

\section{Results on Benchmarks}

\begin{table}[t]
\centering
\begin{small}
\begin{tabular}{lc@{\hskip6pt}cc@{\hskip6pt}cc@{\hskip6pt}c@{\hskip6pt}}
\toprule
\multirow{2}{*}{\textbf{Method}} & \multicolumn{2}{c}{Ans} &
\multicolumn{2}{c}{Sup} &
\multicolumn{2}{c}{Joint} \\ 
\cmidrule(lr){2-3}
\cmidrule(lr){4-5}
\cmidrule(lr){6-7}
& EM & F1 & EM & F1 & EM & F1\\
\midrule
Yang~\shortcite{yang2018hotpotqa} & 24.7 & 34.4 & 5.3 & 41.0 & 2.5 & 17.7\\
Ding~\shortcite{ding2019coggraph} & 37.6 & 49.4 & 23.1 & 58.5 & 12.2 & 35.3 \\
whole pip. & \textbf{46.5} & \textbf{58.8} & \textbf{39.9} & \textbf{71.5} & \textbf{26.6} & \textbf{49.2}\\
\multicolumn{7}{l}{\footnotesize{\textbf{\textit{Dev set}}}}\\
\midrule
Yang~\shortcite{yang2018hotpotqa} & 24.0	& 32.9	& 3.9 & 37.7 &	1.9 & 16.2\\
MUPPET & 30.6 & 40.3 & 16.7 & 47.3 & 10.9 & 27.0\\
Ding~\shortcite{ding2019coggraph} & 37.1 & 48.9 & 22.8 & 57.7 & 12.4 & 34.9\\
whole pip. & \textbf{45.3} & \textbf{57.3} & \textbf{38.7} & \textbf{70.8} & \textbf{25.1} &	\textbf{47.6} \\
\multicolumn{7}{l}{\footnotesize{\textbf{\textit{Test set}}}}\\
\bottomrule
\end{tabular}
\end{small}
\caption{Results of systems on \hpqa.}
\label{tab:hotpot_test}
\end{table}

\begin{table}[t]
\centering
\begin{tabular}{lccc}
\toprule
Model & F1 & LA & FS \\
\midrule
Hanselowski~\shortcite{hanselowski2018ukp_3rd} & - & 68.49 & 64.74\\
Yoneda~\shortcite{yoneda2018ucl_2nd} & 35.84 & 69.66 & 65.41\\
Nie~\shortcite{nie2019combining} & 51.37 & 69.64 & 66.15\\
Full system (single) & \textbf{76.87} & \textbf{75.12} & \textbf{70.18}\\
\multicolumn{4}{l}{\footnotesize{\textbf{\textit{Dev set}}}}\\
\midrule
Hanselowski~\shortcite{hanselowski2018ukp_3rd} & 37.33 & 65.22 & 61.32\\
Yoneda~\shortcite{yoneda2018ucl_2nd} & 35.21 & 67.44 & 62.34\\
Nie~\shortcite{nie2019combining} & 52.81 & 68.16 & 64.23\\
Full system (single) & \textbf{74.62} & \textbf{72.56} & \textbf{67.26} \\
\multicolumn{4}{l}{\footnotesize{\textbf{\textit{Test set}}}}\\
\bottomrule
\end{tabular}
\caption{Performance of systems on \fever. ``F1'' indicates the sentence-level evidence F1 score. ``LA'' indicates Label Acc. without considering the evidence prediction. ``FS''=FEVER Score~\cite{Thorne18Fever}}
\label{tab:fever_test}
\vspace{-5pt}
\end{table}
We chose the best system based on the dev set, and used that for submitting private test predictions on both \fever and \hpqa\footnote{Results can also be found at the leaderboard websites for the two datasets: \url{https://hotpotqa.github.io} and \url{https://competitions.codalab.org/competitions/18814}}.

As can be seen in Table~\ref{tab:hotpot_test}, with the proposed hierarchical system design, the whole pipeline system achieves new start-of-the-art on \hpqa with large-margin improvements on all the metrics. More specifically, the biggest improvement comes from the EM for the supporting fact which in turn leads to doubling of the joint EM on previous best results. The scores for answer predictions are also higher than all previous best results with $\sim$8 absolute points increase on EM and $\sim$9 absolute points on F1. All the improvements are consistent between test and dev set evaluation.

Similarly for \fever, we showed F1 for evidence, the Label Accuracy, and the FEVER Score (same as benchmark evaluation) for models in Table~\ref{tab:fever_test}. Our system obtained substantially higher scores than all previously published results with a $\sim$4 and $\sim$3 points absolute improvement on Label Accuracy and \fever  Score. In particular, the system gains 74.62 on the evidence F1, 22 points greater that of the second system, demonstrating its ability on semantic retrieval.

Previous systems~\cite{ding2019coggraph,yang2018hotpotqa} on \hpqa treat supporting fact retrieval (sentence-level retrieval) just as an auxiliary task for providing extra model explainability. In \newcite{nie2019combining}, although they used a similar three-stage system for \fever, they only applied one neural retrieval module at sentence-level which potentially weaken its retrieval ability. Both of these previous best systems are different from our fully hierarchical pipeline approach. These observations lead to the assumption that the performance gain comes mainly from the hierarchical retrieval and its positive effects on downstream. Therefore, to validate the system design decisions in Sec~\ref{sec:system_setup} and reveal the importance of semantic retrieval towards downstream, we conducted a series of ablation and analysis experiments on all the modules. We started by examining the necessity of both paragraph and sentence retrieval and give insights on why both of them matters.

\section{Analysis and Ablations}
\label{sec:analysis}
Intuitively, both the paragraph-level and sentence-level retrieval sub-module help speeding up the downstream processing. More importantly, since downstream modules were trained by sampled data from upstream modules, both of neural retrieval sub-modules also play an implicit but important role in controlling the immediate retrieval distribution i.e. the distribution of set $\mathbf{P_N}$ and set $\mathbf{S}$ (as shown in Fig.~\ref{fig:fig_system_overview}), and providing better inference data and training data for downstream modules.

\subsection{Ablation Studies} 
\paragraph{Setups:} To reveal the importance of neural retrieval modules at both paragraph and sentence level for maintaining the performance of the overall system, we removed either of them and examine the consequences. Because the removal of a module in the pipeline might change the distribution of the input of the downstream modules, we re-trained all the downstream modules accordingly. To be specific, in the system without the paragraph-level neural retrieval module, we re-trained the sentence-level retrieval module with negative sentences directly sampled from the term-based retrieval set and then also re-trained the downstream QA or verification module. In the system without the sentence-level neural retrieval module, we re-train the downstream QA or verification module by sampling data from both ground truth set and retrieved set directly from the paragraph-level module. We tested the simplified systems on both \fever and \hpqa.

\begin{table*}[t]
\centering
\begin{small}
\begin{tabular}{lccccccccccccc}
\toprule
\multirow{2}{*}{\textbf{Method}} &
\multicolumn{3}{c}{P-Level Retrieval} & \multicolumn{4}{c}{S-Level Retrieval} &
\multicolumn{2}{c}{Answer} &
\multicolumn{2}{c}{Joint} \\
\cmidrule(lr){2-4}
\cmidrule(lr){5-8}
\cmidrule(lr){9-10}
\cmidrule(lr){11-12}
 & Prec. & Rec. & F1 & EM & Prec. & Rec. & F1 & EM & F1 & EM & F1\\
\midrule
Whole Pip. & 35.17 & 87.93 & 50.25 & \textbf{39.86} & \textbf{75.60} & \textbf{71.15} & \textbf{71.54} & \textbf{46.50} & \textbf{58.81} & \textbf{26.60} & \textbf{49.16} \\
Pip. w/o p-level & 6.02 & 89.53 & 11.19 & 0.58 & 29.57 & 60.71 & 38.84 & 31.23 & 41.30 & 0.34 & 19.71 \\
Pip. w/o s-level & 35.17 & 87.92 & 50.25 & - & - & - & - & 44.77 & 56.71 & - & - \\
\bottomrule
\end{tabular}
\end{small}
\vspace{-5pt}
\caption{Ablation over the paragraph-level and sentence-level neural retrieval sub-modules on \hpqa.}
\label{tab:hotpot_p_level_ablation}
\vspace{-5pt}
\end{table*}

\begin{table*}[t]
\centering
\begin{small}
\begin{tabular}{lccccccccccccc}
\toprule
\multirow{2}{*}{\textbf{Method}} &
\multicolumn{4}{c}{P-Level Retrieval} & \multicolumn{4}{c}{S-Level Retrieval} &
\multicolumn{3}{c}{Verification} \\ 
\cmidrule(lr){2-5}
\cmidrule(lr){6-9}
\cmidrule(lr){10-12}
& Orcl. & Prec. & Rec. & F1 & Orcl. & Prec. & Rec. & F1 & LA & FS & \scriptsize{L-F1 (S/R/N)}\\
\midrule
Whole Pip. & 94.15 & 48.84 & 91.23 & 63.62 & 88.92 & \textbf{71.29} & 83.38 & 76.87 & \textbf{70.18} & \textbf{75.01} & \footnotesize{81.7/75.7/\textbf{67.1}}\\
Pip. w/o p-level & \textbf{94.69} & 18.11 & \textbf{92.03} & 30.27 & \textbf{91.07} & 44.47 & 
\textbf{86.60} & 58.77 & 61.55 & 67.01 & \footnotesize{76.5/72.7/40.8}\\
Pip. w/o s-level & 94.15 & 48.84 & 91.23 & 63.62 & - & - & - & - & 55.92 & 61.04 & \footnotesize{72.1/67.6/\underline{27.7}}\\
\bottomrule
\end{tabular}
\end{small}
\vspace{-5pt}
\caption{Ablation over the paragraph-level and sentence-level neural retrieval sub-modules on \fever.
``LA''=Label Accuracy; ``FS''=FEVER Score; ``Orcl.'' is the oracle upperbound of FEVER Score assuming all downstream modules are perfect. ``L-F1 (S/R/N)'' means the classification f1 scores on the three verification labels: \textsc{Support}, \textsc{Refute}, and \textsc{Not Enough Info}.}
\vspace{-10pt}
\label{tab:fever_p_level_ablation}
\end{table*}

\paragraph{Results:} Table~\ref{tab:hotpot_p_level_ablation} and~\ref{tab:fever_p_level_ablation} shows the ablation results for the two neural retrieval modules at both paragraph and sentence level on \hpqa and \fever.
To begin with, we can see that removing paragraph-level retrieval module significantly reduces the precision for sentence-level retrieval and the corresponding F1 on both tasks. More importantly, this loss of retrieval precision also led to substantial decreases for all the downstream scores on both QA and verification task in spite of their higher upper-bound and recall scores. This indicates that the negative effects on downstream module induced by the omission of paragraph-level retrieval can not be amended by the sentence-level retrieval module, and focusing semantic retrieval merely on improving the recall or the upper-bound of final score will risk jeopardizing the performance of the overall system.

Next, the removal of sentence-level retrieval module induces a $\sim$2 point drop on EM and F1 score in the QA task, and a $\sim$15 point drop on FEVER Score in the verification task. This suggests that rather than just enhance explainability for QA, the sentence-level retrieval module can also help pinpoint relevant information and reduce the noise in the evidence that might otherwise distract the downstream comprehension module. Another interesting finding is that without sentence-level retrieval module, the QA module suffered much less than the verification module; 
conversely, the removal of paragraph-level retrieval neural induces a 11 point drop on answer EM comparing to a $\sim$9 point drop on Label Accuracy in the verification task.
This seems to indicate that the downstream QA module relies more on the upstream paragraph-level retrieval whereas the verification module relies more on the upstream sentence-level retrieval.
Finally, we also evaluate the F1 score on \fever for each classification label and we observe a significant drop of F1 on \textsc{Not Enough Info} category without retrieval module, meaning that semantic retrieval is vital for the downstream verification module's discriminative ability on \textsc{Not Enough Info} label.

\subsection{Sub-Module Change Analysis}
To further study the effects of upstream semantic retrieval towards downstream tasks, we change training or inference data between intermediate layers and then examine how this modification will affect the downstream performance.

\begin{figure}[t]
	\centering
    \includegraphics[clip,width=0.45\textwidth]{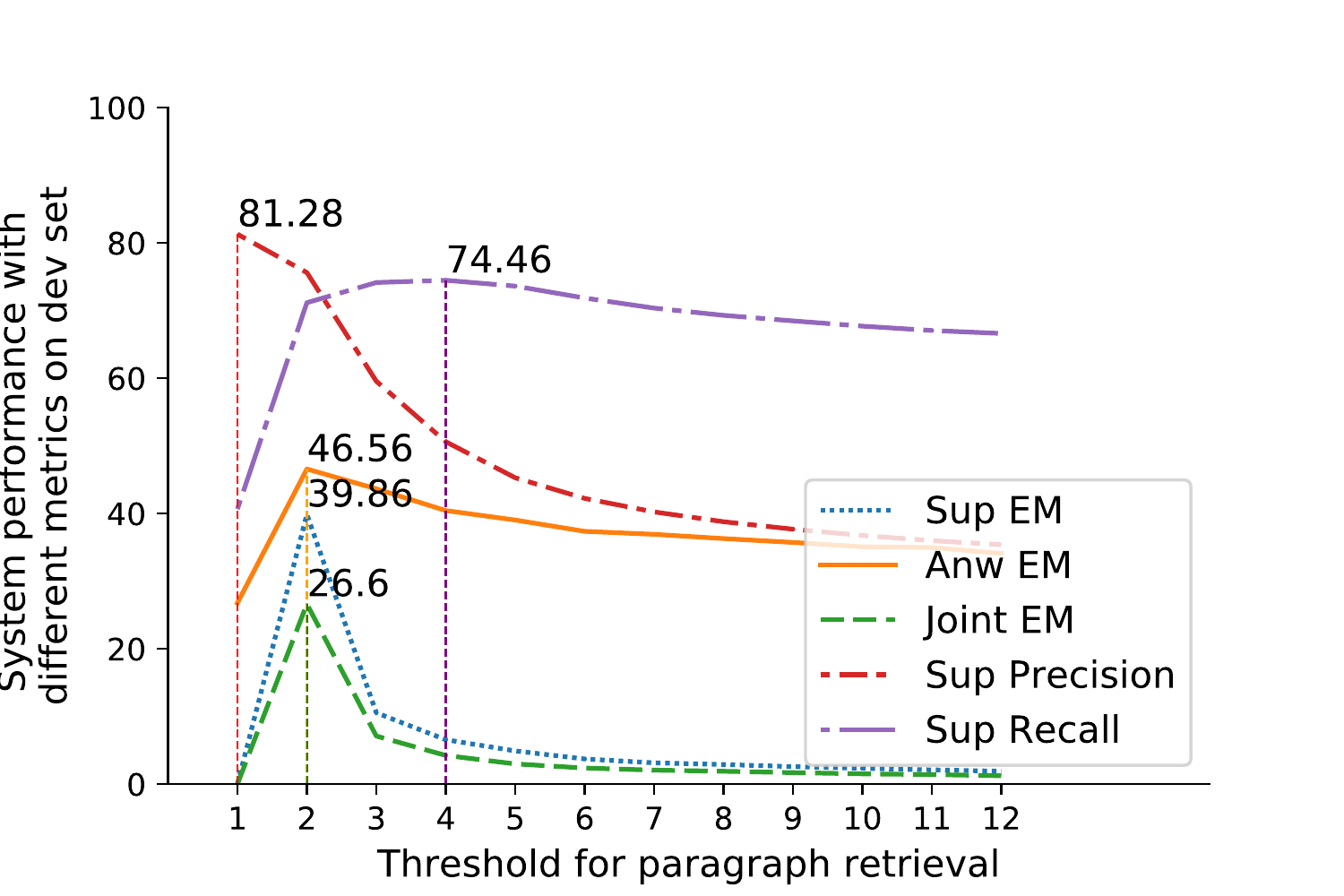}
    \vspace{-7pt}
    \caption{The results of EM for supporting fact, answer prediction and joint score, and the results of supporting fact precision and recall with different values of $k_p$ at paragraph-level retrieval on \hpqa.
    \label{fig:hpqa_p_level_effects}
}
\vspace{-10pt}
\end{figure}

\begin{figure}[t]
	\centering
    \includegraphics[clip,width=0.45\textwidth]{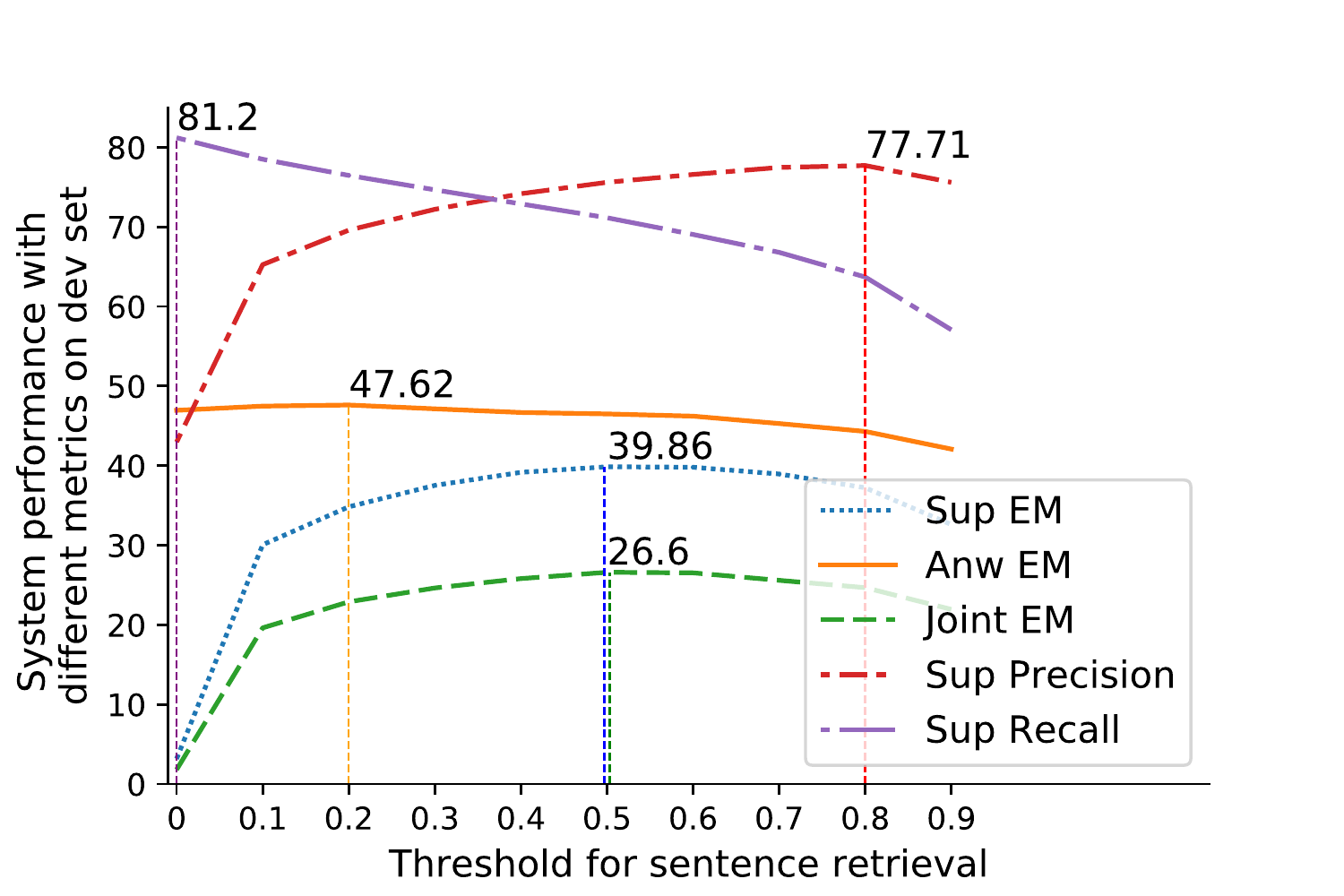}
    \vspace{-5pt}
    \caption{The results of EM for supporting fact, answer prediction and joint score, and the results of supporting fact precision and recall with different values of $h_s$ at sentence-level retrieval on \hpqa.
    \label{fig:hpqa_s_level_effects}}
    \vspace{-5pt}
\end{figure}

\begin{figure}[t]
	\centering
    \includegraphics[clip,width=0.45\textwidth]{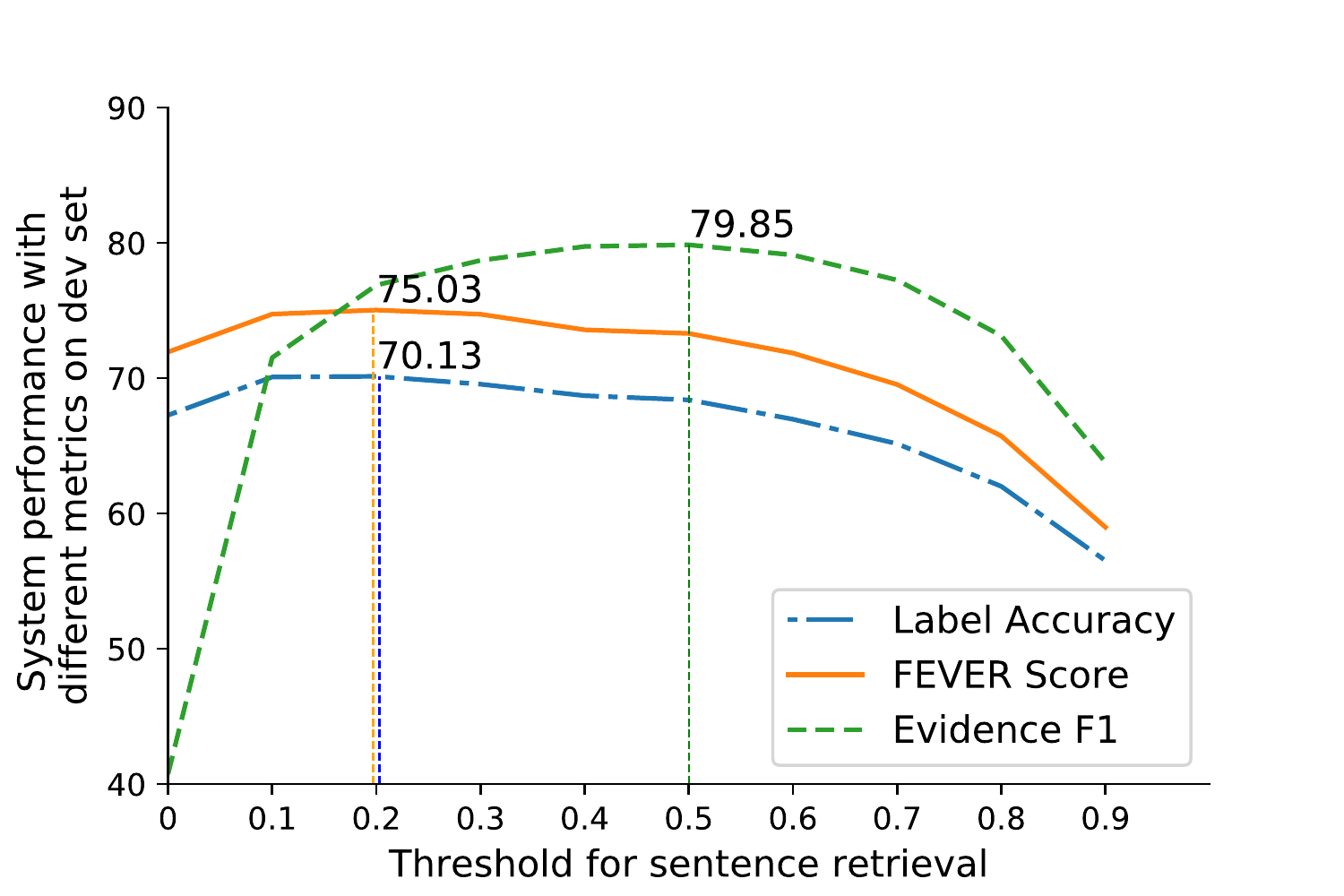}
    \vspace{-5pt}
    \caption{The results of Label Accuracy, FEVER Score, and Evidence F1 with different values of $h_s$ at sentence-level retrieval on \fever.
    \label{fig:fever_s_level_effects}}
    \vspace{-5pt}
\end{figure}

\subsubsection{Effects of Paragraph-level Retrieval}
We fixed $h_p=0$ (the value achieving the best performance) and re-trained all the downstream parameters and track their performance as $k_p$ (the number of selected paragraph) being changed from 1 to 12. The increasing of $k_p$ means a potential higher coverage of the answer but more noise in the retrieved facts.
Fig.~\ref{fig:hpqa_p_level_effects} shows the results. As can be seen that the EM scores for supporting fact retrieval, answer prediction, and joint performance increase sharply when $k_p$ is changed from 1 to 2. This is consistent with the fact that at least two paragraphs are required to ask each question in \hpqa. Then, after the peak, every score decrease as $k_p$ becomes larger except the recall of supporting fact which peaks when $k_p=4$. This indicates that even though the neural sentence-level retrieval module poccesses a certain level of ability to select correct facts from noisier upstream information, the final QA module is more sensitive to upstream data and fails to maintain the overall system performance.
Moreover, the reduction on answer EM and joint EM suggests that it might be risky to give too much information for downstream modules with a unit of a paragraph.

\subsubsection{Effects of Sentence-level Retrieval}
Similarly, to study the effects of neural sentence-level retrieval module towards downstream QA and verification modules, we fixed $k_s$ to be 5 and set $h_s$ ranging from 0.1 to 0.9 with a 0.1 interval.
Then, we re-trained the downstream QA and verification modules with different $h_s$ value and experimented on both \hpqa and \fever.

\noindent \textbf{Question Answering:}
Fig.~\ref{fig:hpqa_s_level_effects} shows the trend of performance.
Intuitively, the precision increase while the recall decrease as the system becomes more strict about the retrieved sentences. The EM score for supporting fact retrieval and joint performance reaches their highest value when $h_s=0.5$, a natural balancing point between precision and recall. More interestingly, the EM score for answer prediction peaks when $h_s=0.2$ and where the recall is higher than the precision. This misalignment between answer prediction performance and retrieval performance indicates that unlike the observation at paragraph-level, the downstream QA module is able to stand a certain amount of noise at sentence-level and benefit from a higher recall.

\noindent \textbf{Fact Verification:}
Fig.~\ref{fig:fever_s_level_effects} shows the trends for Label Accuracy, \fever Score, and Evidence F1 by modifying upstream sentence-level threshold $h_s$. We observed that the general trend is similar to that of QA task where both the label accuracy and \fever score peak at $h_s=0.2$ whereas the retrieval F1 peaks at $h_s=0.5$. Note that, although the downstream verification could take advantage of a higher recall, the module is more sensitive to sentence-level retrieval comparing to the QA module in \hpqa. More detailed results are in the Appendix.

\subsection{Answer Breakdown}
\begin{figure}[t]
	\centering
    \includegraphics[clip,width=0.45\textwidth]{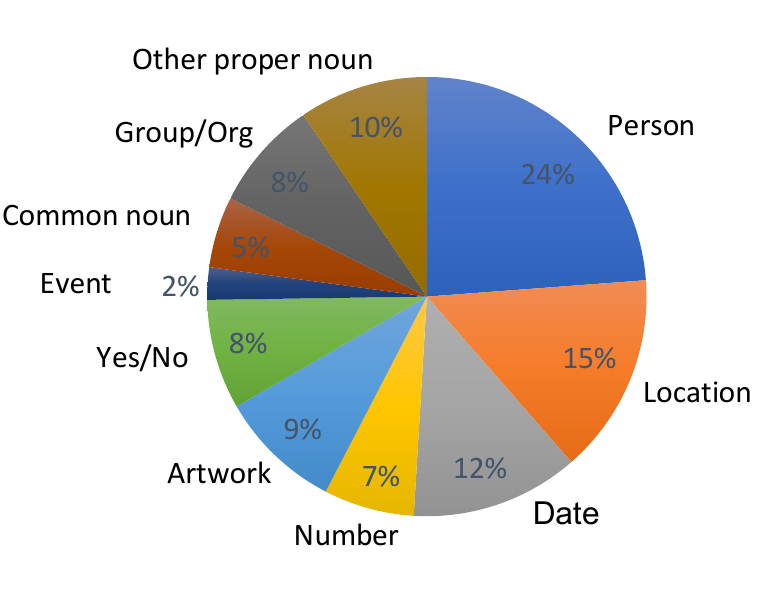}
    \vspace{-10pt}
    \caption{Proportion of answer types.
    \label{fig:answer_type_breakdown}
}
\end{figure}

We further sample 200 examples from \hpqa and manually tag them according to several common answer types~\cite{yang2018hotpotqa}. The proportion of different answer types is shown in Figure~\ref{fig:answer_type_breakdown}. The performance of the system on each answer type is shown in Table~\ref{tab:answer_breakdown_performance}. The most frequent answer type is 'Person' (24\%) and the least frequent answer type is 'Event' (2\%). It is also interesting to note that the model performs the best in Yes/No questions as shown in Table~\ref{tab:answer_breakdown_performance}, reaching an accuracy of 70.6\%.

\begin{table}[t]
\centering
\begin{tabular}{lrrr}
\toprule
Answer Type & Total & Correct & Acc. (\%)\\
\midrule
Person & 50 & 28 & 56.0 \\
Location & 31 & 14 & 45.2 \\
Date & 26 & 13 & 50.0 \\
Number & 14 & 4 & 28.6 \\
Artwork & 19 & 7 & 36.8 \\
Yes/No & 17 & 12 & \bf{70.6} \\
Event & 5 & 2 & 40.0 \\
Common noun & 11 & 3 & 27.3 \\
Group/Org & 17 & 6 & 35.3 \\
Other PN & 20 & 9 & 45.0 \\
\midrule
Total & 200 & 98 & 49.0 \\
\bottomrule
\end{tabular}
\caption{System performance on different answer types. ``PN''= Proper Noun}
\label{tab:answer_breakdown_performance}
\vspace{-5pt}
\end{table}

\subsection{Examples}
Fig.~\ref{fig:example} shows an example that is correctly handled by the full pipeline system but not by the system without paragraph-level retrieval module. We can see that it is very difficult to filter the distracting sentence after sentence-level either by the sentence retrieval module or the QA module.

Above findings in both \fever and \hpqa bring us some important guidelines for MRS:
(1) A paragraph-level retrieval module is imperative;
(2) Downstream task module is able to undertake a certain amount of noise from sentence-level retrieval;
(3) Cascade effects on downstream task might be caused by modification at paragraph-level retrieval.

\begin{figure}[t]
\fbox{
\begin{minipage}{0.90\linewidth}
\begin{small}
\textit{\textbf{Question:}} Wojtek Wolski played for what team based in the Miami metropolitan area?\\
\textit{\textbf{GT Answer:}} Florida Panthers \vspace{3pt}

\textit{\textbf{GT Facts:}}\\
{[}\textbf{Florida Panthers,0}{]}: The Florida Panthers are a professional ice hockey team based in the Miami metropolitan area. $(\textit{P-Score}: 0.99; \textit{S-Score}: 0.98)$\\
{[}\textbf{Wojtek Wolski,1}{]}: In the NHL, he has played for the Colorado Avalanche, Phoenix Coyotes, New York Rangers, \ul{Florida Panthers}, and the Washington Capitals. $(\textit{P-Score}: 0.98; \textit{S-Score}: 0.95)$\vspace{3pt}

\textit{\textbf{Distracting Fact:}}\\
{[}\textbf{History of the Miami Dolphins,0}{]}:
\dashuline{The Miami Dolphins} are a professional American football franchise based in the Miami metropolitan area. $(\textit{P-Score}: 0.56; \textit{S-Score}: 0.97)$\vspace{3pt}

\textit{\textbf{Wrong Answer :}} The Miami Dolphins
\end{small}
\end{minipage}}
\vspace{-5pt}
\caption{An example with a distracting fact. $\text{P-Score}$ and $\text{S-Score}$ are the retrieval score at paragraph and sentence level respectively. The full pipeline was able to filter the distracting fact and give the correct answer. The wrong answer in the figure was produced by the system without paragraph-level retrieval module.}
\label{fig:example}
\vspace{-10pt}
\end{figure}

\section{Conclusion}
We proposed a simple yet effective hierarchical pipeline system that achieves state-of-the-art results on two MRS tasks. Ablation studies demonstrate the importance of semantic retrieval at both paragraph and sentence levels in the MRS system. The work can give general guidelines on MRS modeling and inspire future research on the relationship between semantic retrieval and downstream comprehension in a joint setting.

\section*{Acknowledgments}
We thank the reviewers for their helpful comments and Yicheng Wang for useful comments. This work was supported by awards from Verisk, Google, Facebook, Salesforce, and Adobe (plus Amazon and Google GPU cloud credits). The views, opinions, and/or findings contained in this article are those of the authors and should not be interpreted as representing the official views or policies, either expressed or implied, of the funding agency.

\bibliography{emnlp-ijcnlp-2019}

\begin{thebibliography}{25}
\expandafter\ifx\csname natexlab\endcsname\relax\def\natexlab#1{#1}\fi

\bibitem[{Bowman et~al.(2015)Bowman, Angeli, Potts, and
  Manning}]{snli:emnlp2015}
Samuel~R. Bowman, Gabor Angeli, Christopher Potts, and Christopher~D. Manning.
  2015.
\newblock A large annotated corpus for learning natural language inference.
\newblock In \emph{Conference on Empirical Methods in Natural Language
  Processing (EMNLP)}. Association for Computational Linguistics.

\bibitem[{Chen et~al.(2017)Chen, Fisch, Weston, and Bordes}]{chen2017drqa}
Danqi Chen, Adam Fisch, Jason Weston, and Antoine Bordes. 2017.
\newblock Reading {Wikipedia} to answer open-domain questions.
\newblock In \emph{Association for Computational Linguistics (ACL)}.

\bibitem[{Clark and Gardner(2017)}]{clark2017simple}
Christopher Clark and Matt Gardner. 2017.
\newblock Simple and effective multi-paragraph reading comprehension.
\newblock \emph{Association for Computational Linguistics ({ACL})}.

\bibitem[{Dehghani et~al.(2017)Dehghani, Zamani, Severyn, Kamps, and
  Croft}]{dehghani2017neural}
Mostafa Dehghani, Hamed Zamani, Aliaksei Severyn, Jaap Kamps, and W~Bruce
  Croft. 2017.
\newblock Neural ranking models with weak supervision.
\newblock In \emph{SIGIR}.

\bibitem[{Devlin et~al.(2018)Devlin, Chang, Lee, and
  Toutanova}]{devlin2018bert}
Jacob Devlin, Ming-Wei Chang, Kenton Lee, and Kristina Toutanova. 2018.
\newblock Bert: Pre-training of deep bidirectional transformers for language
  understanding.
\newblock \emph{arXiv preprint arXiv:1810.04805}.

\bibitem[{Ding et~al.(2019)Ding, Zhou, Chen, Yang, and Tang}]{ding2019coggraph}
Ming Ding, Chang Zhou, Qibin Chen, Hongxia Yang, and Jie Tang. 2019.
\newblock Cognitive graph for multi-hop reading comprehension at scale.
\newblock In \emph{Association for Computational Linguistics ({ACL})}.

\bibitem[{Guo et~al.(2016)Guo, Fan, Ai, and Croft}]{guo2016deep}
Jiafeng Guo, Yixing Fan, Qingyao Ai, and W~Bruce Croft. 2016.
\newblock A deep relevance matching model for ad-hoc retrieval.
\newblock In \emph{CIKM}.

\bibitem[{Hanselowski et~al.(2018)Hanselowski, Zhang, Li, Sorokin, Schiller,
  Schulz, and Gurevych}]{hanselowski2018ukp_3rd}
Andreas Hanselowski, Hao Zhang, Zile Li, Daniil Sorokin, Benjamin Schiller,
  Claudia Schulz, and Iryna Gurevych. 2018.
\newblock Ukp-athene: Multi-sentence textual entailment for claim verification.
\newblock In \emph{The 1st Workshop on Fact Extraction and Verification}.

\bibitem[{Htut et~al.(2018)Htut, Bowman, and Cho}]{htut2018training}
Phu~Mon Htut, Samuel~R Bowman, and Kyunghyun Cho. 2018.
\newblock Training a ranking function for open-domain question answering.
\newblock In \emph{NAACL-HLT}.

\bibitem[{Huang et~al.(2013)Huang, He, Gao, Deng, Acero, and
  Heck}]{huang2013learning}
Po-Sen Huang, Xiaodong He, Jianfeng Gao, Li~Deng, Alex Acero, and Larry Heck.
  2013.
\newblock Learning deep structured semantic models for web search using
  clickthrough data.
\newblock In \emph{CIKM}.

\bibitem[{Ioffe and Szegedy(2015)}]{ioffe2015batchnorm}
Sergey Ioffe and Christian Szegedy. 2015.
\newblock Batch normalization: Accelerating deep network training by reducing
  internal covariate shift.
\newblock In \emph{International Conference on International Conference on
  Machine Learning ({ICML})}.

\bibitem[{Joshi et~al.(2017)Joshi, Choi, Weld, and
  Zettlemoyer}]{Joshi2017trivia_qa}
Mandar Joshi, Eunsol Choi, Daniel~S. Weld, and Luke Zettlemoyer. 2017.
\newblock Triviaqa: A large scale distantly supervised challenge dataset for
  reading comprehension.
\newblock In \emph{Proceedings of the 55th Annual Meeting of the Association
  for Computational Linguistics}, Vancouver, Canada. Association for
  Computational Linguistics.

\bibitem[{Lee et~al.(2018)Lee, Yun, Kim, Ko, and Kang}]{lee2018ranking}
Jinhyuk Lee, Seongjun Yun, Hyunjae Kim, Miyoung Ko, and Jaewoo Kang. 2018.
\newblock Ranking paragraphs for improving answer recall in open-domain
  question answering.
\newblock In \emph{Conference on Empirical Methods in Natural Language
  Processing ({EMNLP})}.

\bibitem[{MacCartney and Manning(2009)}]{maccartney2009NLI}
Bill MacCartney and Christopher~D Manning. 2009.
\newblock \emph{Natural language inference}.
\newblock Citeseer.

\bibitem[{Mitra et~al.(2017)Mitra, Diaz, and Craswell}]{mitra2017learning}
Bhaskar Mitra, Fernando Diaz, and Nick Craswell. 2017.
\newblock Learning to match using local and distributed representations of text
  for web search.
\newblock In \emph{WWW}.

\bibitem[{Nguyen et~al.(2016)Nguyen, Rosenberg, Song, Gao, Tiwary, Majumder,
  and Deng}]{nguyen2016ms_marco}
Tri Nguyen, Mir Rosenberg, Xia Song, Jianfeng Gao, Saurabh Tiwary, Rangan
  Majumder, and Li~Deng. 2016.
\newblock Ms marco: A human generated machine reading comprehension dataset.
\newblock \emph{arXiv preprint arXiv:1611.09268}.

\bibitem[{Nie et~al.(2019)Nie, Chen, and Bansal}]{nie2019combining}
Yixin Nie, Haonan Chen, and Mohit Bansal. 2019.
\newblock Combining fact extraction and verification with neural semantic
  matching networks.
\newblock In \emph{Association for the Advancement of Artificial Intelligence
  ({AAAI})}.

\bibitem[{Peters et~al.(2018)Peters, Neumann, Iyyer, Gardner, Clark, Lee, and
  Zettlemoyer}]{peters2018ELMo}
Matthew~E Peters, Mark Neumann, Mohit Iyyer, Matt Gardner, Christopher Clark,
  Kenton Lee, and Luke Zettlemoyer. 2018.
\newblock Deep contextualized word representations.
\newblock \emph{NAACL-HLT}.

\bibitem[{Radford et~al.(2018)Radford, Narasimhan, Salimans, and
  Sutskever}]{radford2018gpt_1}
Alec Radford, Karthik Narasimhan, Tim Salimans, and Ilya Sutskever. 2018.
\newblock Improving language understanding by generative pre-training.
\newblock \emph{URL https://s3-us-west-2. amazonaws.
  com/openai-assets/research-covers/languageunsupervised/language understanding
  paper. pdf}.

\bibitem[{Thorne and Vlachos(2018)}]{thorne2018automated_survey}
James Thorne and Andreas Vlachos. 2018.
\newblock Automated fact checking: Task formulations, methods and future
  directions.
\newblock In \emph{International Conference on Computational Linguistics
  ({COLIN})}.

\bibitem[{Thorne et~al.(2018)Thorne, Vlachos, Christodoulopoulos, and
  Mittal}]{Thorne18Fever}
James Thorne, Andreas Vlachos, Christos Christodoulopoulos, and Arpit Mittal.
  2018.
\newblock {FEVER}: a large-scale dataset for fact extraction and
  {VERification}.
\newblock In \emph{NAACL-HLT}.

\bibitem[{Wang et~al.(2018)Wang, Yu, Guo, Wang, Klinger, Zhang, Chang, Tesauro,
  Zhou, and Jiang}]{wang2018r3}
Shuohang Wang, Mo~Yu, Xiaoxiao Guo, Zhiguo Wang, Tim Klinger, Wei Zhang, Shiyu
  Chang, Gerry Tesauro, Bowen Zhou, and Jing Jiang. 2018.
\newblock R: Reinforced ranker-reader for open-domain question answering.
\newblock In \emph{Thirty-Second AAAI Conference on Artificial Intelligence}.

\bibitem[{Welbl et~al.(2018)Welbl, Stenetorp, and
  Riedel}]{welbl2018constructing}
Johannes Welbl, Pontus Stenetorp, and Sebastian Riedel. 2018.
\newblock Constructing datasets for multi-hop reading comprehension across
  documents.
\newblock \emph{Transactions of the Association of Computational Linguistics},
  6:287--302.

\bibitem[{Yang et~al.(2018)Yang, Qi, Zhang, Bengio, Cohen, Salakhutdinov, and
  Manning}]{yang2018hotpotqa}
Zhilin Yang, Peng Qi, Saizheng Zhang, Yoshua Bengio, William~W Cohen, Ruslan
  Salakhutdinov, and Christopher~D Manning. 2018.
\newblock Hotpotqa: A dataset for diverse, explainable multi-hop question
  answering.
\newblock In \emph{Conference on Empirical Methods in Natural Language
  Processing ({EMNLP})}.

\bibitem[{Yoneda et~al.(2018)Yoneda, Mitchell, Welbl, Stenetorp, and
  Riedel}]{yoneda2018ucl_2nd}
Takuma Yoneda, Jeff Mitchell, Johannes Welbl, Pontus Stenetorp, and Sebastian
  Riedel. 2018.
\newblock Ucl machine reading group: Four factor framework for fact finding
  (hexaf).
\newblock In \emph{The 1st Workshop on Fact Extraction and Verification}.

\end{thebibliography}
\bibliographystyle{acl_natbib}

\section*{Appendix}
\appendix
\section{Training Details}
\label{sec:training_appendix}

\begin{table}[h]
\centering
\begin{small}
\scalebox{0.9}{
\begin{tabular}{lcccc}
\toprule
Module & \footnotesize{BS.} & \footnotesize{\# E.} & $k$ & $h$ \\
\midrule
\footnotesize{P-Level Retri.} & 64 & 3 & \{2, 5\} & \{5e-3, 0.01, 0.1, 0.5\}\\
\footnotesize{S-Level Retri.} & 128 & 3 & \{2, 5\} & [0.1-0.5]\\
\footnotesize{QA} & 32 & 5 & - & - \\
\footnotesize{Verification} & 32 & 5 & - & -\\
\bottomrule
\end{tabular}}
\end{small}
\caption{Hyper-parameter selection for the full pipeline system. $h$ and $k$ are the retrieval filtering hyper-parameters mentioned in the main paper. P-level and S-level indicate paragraph-level and sentence-level respectively. ``\{\}" means values enumerated from a set. ``[]" means values enumerated from a range with interval=0.1 ``BS."=Batch Size ``\# E."=Number of Epochs}
\label{tab:hyper_parameter_pipeline_system}
\end{table}

\begin{table*}[t]
\centering
\begin{tabular}{ccccccccccc}
\toprule
\multirow{2}{*}{$h_p$} &
\multicolumn{4}{c}{S-Level Retrieval} &
\multicolumn{2}{c}{Answer} &
\multicolumn{4}{c}{Joint} \\
\cmidrule(lr){2-5}
\cmidrule(lr){6-7}
\cmidrule(lr){8-11}
 & EM & Prec. & Rec. & F1 & EM & F1 & EM & F1 & Prec. & Rec.\\
\midrule
0 & 3.17 &  42.97  & \textbf{81.20} & 55.03 & 46.95 & 59.73 & 1.81  & 37.42 & 29.99 & \textbf{57.66} \\
0.1 & 30.06 & 65.26 & 78.50 & 69.72 & 47.48 & 59.78 & 19.62 & 47.86 & 46.09 & 55.63\\
0.2 & 34.83 & 69.59 & 76.47 & 71.28 & \textbf{47.62} & \textbf{59.93} & 22.89 & 49.15 & 49.24 & 54.53\\
0.3 & 37.52 & 72.21 & 74.66 & 71.81 & 47.14 & 59.51 & 24.63 & \textbf{49.44} & 50.67 & 53.55\\
0.4 & 39.16 & 74.17 & 72.89 & \textbf{71.87} & 46.68 & 58.96 & 25.81 & 49.18 & 51.62 & 51.90\\
0.5 & \textbf{39.86} & 75.60 & 71.15 & 71.54 & 46.50 & 58.81 & \textbf{26.60} & 49.16 & 52.59 & 50.83\\
0.6 & 39.80 & 76.59 & 69.05 & 70.72 & 46.22 & 58.31 & 26.53 & 48.48 & 52.86 & 49.48\\
0.7 & 38.95 & 77.47 & 66.80 & 69.71 & 45.29 & 57.47 & 25.96 & 47.59 & \textbf{53.06} & 47.86\\
0.8 & 37.22 & \textbf{77.71} & 63.70 & 67.78 & 44.30 & 55.99 & 24.67 & 45.92 & 52.41 & 45.32\\
0.9 & 32.60 & 75.60 & 57.07 & 62.69 & 42.08 & 52.85 & 21.44 & 42.26 & 50.48 & 40.61\\
\bottomrule
\end{tabular}
\caption{Detailed Results of downstream sentence-level retrieval and question answering with different values of $h_s$ on \hpqa.}
\label{tab:hotpot_detail_s_level_results}
\end{table*}

The hyper-parameters were chosen based on the performance of the system on the dev set. The hyper-parameters search space is shown in Table~\ref{tab:hyper_parameter_pipeline_system} and the learning rate was set to $10^{-5}$ in all experiments.

\section{Term-Based Retrieval Details}

\paragraph{FEVER}
We used the same key-word matching method in \newcite{nie2019combining} to get a candidate set for each query. We also used TF-IDF~\cite{chen2017drqa} method to get top-5 related documents for each query. Then, the two sets were combined to get final term-based retrieval set for FEVER. The mean and standard deviation of the number of the retrieved paragraph in the merged set were 8.06 and 4.88.

\paragraph{\hpqa}
We first used the same procedure on \fever to get an initial candidate set for each query in \hpqa. Because \hpqa requires at least 2-hop reasoning for each query, we then extract all the hyperlinked documents from the retrieved documents in the initial candidate set, rank them with TF-IDF~\cite{chen2017drqa} score and then select top-5 most related documents and add them to the candidate set. This gives the final term-based retrieval set for \hpqa. The mean and standard deviation of the number of the retrieved paragraph for each query in \hpqa were 39.43 and 16.05.

\section{Detailed Results} 
\begin{itemize}
    \item The results of sentence-level retrieval and downstream QA with different values of $h_s$ on \hpqa are in Table~\ref{tab:hotpot_detail_s_level_results}.
    \item The results of sentence-level retrieval and downstream verification with different values of $h_s$ on \fever are in Table~\ref{tab:fever_detail_s_level_results}.
    \item The results of sentence-level retrieval and downstream QA with different values of $k_p$ on \hpqa are in Table~\ref{tab:hotpot_detail_p_level_results}.
\end{itemize}

\begin{table*}[t]
\centering
\begin{tabular}{cccccc}
\toprule
\multirow{2}{*}{$h_s$} &
\multicolumn{3}{c}{S-Level Retrieval} &
\multicolumn{2}{c}{Verification} \\ 
\cmidrule(lr){2-4}
\cmidrule(lr){5-6}
& Precision & Recall & F1 & Label Accuracy & FEVER Score\\
\midrule

0   & 26.38 & \textbf{89.28} & 40.74 & 71.93 & 67.26\\
0.1 & 61.41 & 86.27 & 71.51 & 74.73 & 70.08\\
0.2 & 71.29 & 83.38 & 76.87 & \textbf{75.03} & \textbf{70.13}\\
0.3 & 76.45 & 81.08 & 78.70 & 74.73 & 69.55\\
0.4 & 80.52 & 78.95 & 79.73 & 73.57 & 68.70\\
0.5 & 83.76 & 76.31 & \textbf{79.85} & 73.30 & 68.39\\
0.6 & 86.73 & 72.70 & 79.10 & 71.85 & 66.96\\
0.7 & 89.91 & 67.71 & 77.25 & 69.53 & 65.15\\
0.8 & 93.22 & 60.16 & 73.13 & 65.73 & 62.00\\
0.9 & \textbf{96.52} & 47.63 & 63.78 & 58.98 & 56.51\\
\bottomrule
\end{tabular}
\caption{Results with different $h_s$ on \fever.}
\label{tab:fever_detail_s_level_results}
\end{table*}

% k_p
\begin{table*}[t]
\centering
\begin{tabular}{ccccccccccc}
\toprule
\multirow{2}{*}{$k_p$} &
\multicolumn{4}{c}{S-Level Retrieval} &
\multicolumn{2}{c}{Answer} &
\multicolumn{4}{c}{Joint} \\
\cmidrule(lr){2-5}
\cmidrule(lr){6-7}
\cmidrule(lr){8-11}
 & EM & Prec. & Rec. & F1 & EM & F1 & EM & F1 & Prec. & Rec.\\
\midrule

1 & 0 & \textbf{81.28} & 40.64 & 53.16 & 26.77 & 35.76 & 0 & 21.54 & 33.41 & 17.32\\
2 & \textbf{39.86} & 75.60 & 71.15 & \textbf{71.54} & \textbf{46.56} & \textbf{58.74} & \textbf{26.6} & \textbf{49.09} & \textbf{52.54} & \textbf{50.79}\\
3 & 10.53 & 59.54 & 74.13 & 63.92 & 43.63 & 55.51 & 7.09 & 40.61 & 38.52 & 49.20\\
4 & 6.55 & 50.60 & \textbf{74.46} & 57.98 & 40.42 & 51.99 & 4.22 & 34.57 & 30.84 & 45.98\\
5 & 4.89 & 45.27 & 73.60 & 53.76 & 39.02 & 50.36 & 2.97 & 32.14 & 26.92 & 43.89\\
6 & 3.70 & 42.22 & 71.84 & 51.04 & 37.35 & 48.41 & 2.36 & 28.66 & 24.37 & 41.63\\
7 & 3.13 & 40.22 & 70.35 & 49.15 & 36.91 & 47.70 & 2.05 & 27.49 & 23.10 & 40.60\\
8 & 2.88 & 38.77 & 69.28 & 47.83 & 36.28 & 46.99 & 1.88 & 26.58 & 22.13 & 39.85\\
9 & 2.57 & 37.67 & 68.46 & 46.81 & 35.71 & 46.30 & 1.68 & 25.77 & 21.32 & 38.87\\
10 & 2.31 & 36.74 & 67.68 & 45.94 & 35.07 & 45.74 & 1.50 & 25.05 & 20.56 & 38.21\\
11 & 2.09 & 35.97 & 67.04 & 45.21 & 34.96 & 45.56 & 1.39 & 24.65 & 20.18 & 37.60\\
12 & 1.89 & 35.37 & 66.60 & 44.67 & 34.09 & 44.74 & 1.22 & 23.99 & 19.57 & 36.98\\

\bottomrule
\end{tabular}

\caption{Detailed Results of downstream sentence-level retrieval and question answering with different values of $k_p$ on \hpqa.}
\label{tab:hotpot_detail_p_level_results}
\end{table*}

\section{Examples and Case Study}
We further provide examples, case study and error analysis for the full pipeline system. The examples are shown from Tables~\ref{table:ranker_analysis_1},~\ref{table:ranker_analysis_2},~\ref{table:ranker_analysis_3},~\ref{table:ranker_analysis_4},~\ref{table:ranker_analysis_5}. The examples show high diversity on the semantic level and the error occurs often due to the system's failure of extracting precise (either wrong, surplus or insufficient) information from KB.

\begin{table*}[t]
\centering
\small
\begin{tabular}{rl}
\toprule
Question: & \textit{D1NZ is a series based on what oversteering technique?}\\
\midrule

Ground Truth Facts: & (\textit{D1NZ, 0}) \textit{\underline{\textbf{D1NZ}} is a production car drifting series in New Zealand.}\\
 & (\textit{Drifting (motorsport), 0}) \textit{\underline{\textbf{Drifting}} is a driving technique where the driver}\\ &   \textit{intentionally oversteers...}\\
Ground Truth Answer: & \textit{Drifting}\\
\midrule
Predicted Facts: & (\textit{D1NZ, 0}) \textit{\underline{\textbf{D1NZ}} is a production car drifting series in New Zealand.}\\
 & (\textit{Drifting (motorsport), 0}) \textit{\underline{\textbf{Drifting}} is a driving technique where the driver}\\ &   \textit{intentionally oversteers...}\\
Predicted Answer: & \textit{Drifting}\\
\bottomrule
\end{tabular}
\caption{HotpotQA correct prediction with sufficient evidence.}
\label{tab:hotpot_example_1}
\end{table*}

\begin{table*}[t]
\centering
\small
\begin{tabular}{rl}
\toprule
Question: & \textit{The football manager who recruited David Beckham managed Manchester United} \\
& during what timeframe?\\
\midrule

Ground Truth Facts: & (\textit{1995-96 Manchester United F.C. season, 2}) \textit{Alex Ferguson had sold experienced}\\ 
& \textit{players Paul Ince , Mark Hughes and Andrei Kanchelskis...}\\
 & (\textit{1995-96 Manchester United F.C. season,3}) \textit{Instead , he had drafted in young}\\
 & \textit{players like Nicky Butt, David Beckham, Paul Scholes...}\\
 & (\textit{Alex Ferguson,0}) \textit{Sir Alexander Chapman Ferguson... who managed Manchester }\\
& \textit{United \underline{\textbf{from 1986 to 2013}}.}\\
Ground Truth Answer: & \textit{from 1986 to 2013}\\
\midrule
Predicted Facts: & (\textit{Matt Busby,0}) \textit{Sir Alexander Matthew Busby , CBE... who managed Manchester}\\ & \textit{United between \underline{\textbf{{}1945 and 1969}}...}\\

Predicted Answer: & \textit{1945 and 1969}\\
\bottomrule
\end{tabular}
\caption{HotpotQA incorrect prediction with insufficient/wrong evidence.}
\label{table:ranker_analysis_1}
\end{table*}

\begin{table*}[t]
\centering
\small
\begin{tabular}{rl}
\toprule
Question: & \textit{Where did Ian Harland study prior to studying at the oldest college at the University}\\ & \textit{of Cambridge?}\\
\midrule

Ground Truth Facts: & (\textit{Ian Harland,0}) \textit{From a clerical family... Harland was educated at}\\ &  \textit{\underline{\textbf{The Dragon School in Oxford} and Haileybury.}}\\
 & (\textit{Ian Harland,1}) \textit{He then went to university at Peterhouse, Cambridge, taking a law}\\ &  \textit{degree.}\\
 & (\textit{Peterhouse, Cambridge, 1}) \textit{It is the oldest college of the university...}\\
 
Ground Truth Answer: & \textit{The Dragon School in Oxford}\\
\midrule
Predicted Facts: & (\textit{Ian Harland,0}) \textit{From a clerical family... Harland was educated at}\\ &  \textit{\underline{\textbf{The Dragon School in Oxford}} and Haileybury.}\\
 & (\textit{Ian Harland,1}) \textit{He then went to university at Peterhouse, Cambridge, taking a law}\\ &  \textit{degree.}\\
 & (\textit{Peterhouse, Cambridge, 1}) \textit{It is the oldest college of the university...}\\
  & (\textit{Ian Harland, 2}) \textit{After two years as a schoolmaster at Sunningdale School he studied}\\ &  \textit{for the priesthood at \underline{\textbf{Wycliffe Hall, Oxford}} and be}\\
Predicted Answer: & \textit{Wycliffe Hall , Oxford}\\
\bottomrule
\end{tabular}
\caption{HotpotQA incorrect prediction caused by extra incorrect information.}
\label{table:ranker_analysis_2}
\end{table*}

\begin{table*}[t]
\centering
\small
\begin{tabular}{rl}
\toprule
Claim: & \textit{Heroes' first season had 12 episodes.}\\
\midrule

Ground Truth Facts: & (\textit{Heroes (U.S. TV series), 8}) \textit{The critically acclaimed first season had a run of 23 episodes}\\ &  \textit{and garnered an average of 14.3 million viewers in the United States, receiving the highest} \\ &  \textit{rating for an NBC drama premiere in five years.}\\
Ground Truth Label: & \textit{REFUTES}\\
\midrule
Predicted Facts: & (\textit{Heroes (U.S. TV series), 8}) \textit{The critically acclaimed first season had a run of 23 episodes}\\ &  \textit{and garnered an average of 14.3 million viewers in the United States, receiving the highest} \\ &  \textit{rating for an NBC drama premiere in five years.}\\
Predicted Label: & \textit{REFUTES}\\

\bottomrule
\end{tabular}
\caption{FEVER correct prediction with sufficient evidence}
\label{table:ranker_analysis_3}
\end{table*}

\begin{table*}[t]
\centering
\small
\begin{tabular}{rl}
\toprule
Claim: & \textit{Azithromycin is available as a generic curtain.}\\
\midrule

Ground Truth Facts: & (\textit{Azithromycin, 17}) \textit{Azithromycin is an antibiotic useful for the treatment of a number of }\\ &  \textit{bacterial infections.}\\
& (\textit{Azithromycin, 11}) \textit{Azithromycin is an azalide , a type of macrolide antibiotic.}\\
& (\textit{Azithromycin, 0}) \textit{It is available as a generic medication and is sold under many trade names}\\ &  \textit{worldwide.}\\
Ground Truth Label: & \textit{REFUTES}\\
\midrule
Predicted Facts: & (\textit{Azithromycin, 17}) \textit{It is available as a generic medication and is sold under many trade names}\\ &  \textit{worldwide.}\\
Predicted Label: & \textit{NOT ENOUGH INFO}\\

\bottomrule
\end{tabular}
\caption{FEVER incorrect prediction due to insufficient evidence}
\label{table:ranker_analysis_4}
\end{table*}

\begin{table*}[t]
\centering
\small
\begin{tabular}{rl}
\toprule
Claim: & \textit{University of Chicago Law School is ranked fourth by Brain Leiter on the "Top 15 Schools" }\\ &   \textit{From Which the Most `Prestigious\' Law Firms Hire New Lawyers."}\\
\midrule

Ground Truth Facts: & (\textit{University of Chicago Law School,6}) \textit{Chicago is ranked second by Brian Leiter of the}\\ & \textit{University of Chicago Law School on the Top 15 Schools From Which the Most Prestigious} \\ & \textit{Law Firms Hire New Lawyers, and first for Faculty quality based on American Academy of}  \\ & \textit{Arts and Sciences Membership.}\\

Ground Truth Label: & \textit{REFUTES}\\
\midrule
Predicted Facts: & (\textit{University of Chicago Law School,6}) \textit{Chicago is ranked second by Brian Leiter of the}\\ & \textit{University of Chicago Law School on the Top 15 Schools From Which the Most Prestigious} \\ & \textit{Law Firms Hire New Lawyers, and first for Faculty quality based on American Academy of}  \\ & \textit{Arts and Sciences Membership.}\\
& (\textit{University of Chicago Law School,4}) \textit{U.S. News and World Report ranks Chicago fourth}\\ & \textit{among U.S. law schools, and it is noted for its influence on the economic analysis of law.}\\

Predicted Label: & \textit{NOT ENOUGH INFO}\\

\bottomrule
\end{tabular}
\caption{FEVER incorrect prediction due to extra wrong evidence}
\label{table:ranker_analysis_5}
\end{table*}

\end{document}